\documentclass[preprint,12pt]{elsarticle}

\usepackage{graphicx}
\usepackage[utf8]{inputenc}
\usepackage{lscape}
\usepackage{color} 
\usepackage{url}
\usepackage{amsmath}

\usepackage{latexsym} 
\usepackage{amsmath} % assumes amsmath package installed
\usepackage{amssymb}  % assumes amsmath package installed
\usepackage{amsthm}
\usepackage[normalem]{ulem}
\usepackage{graphicx}
\usepackage{float}
\usepackage{algorithm}
\usepackage{algpseudocode}
\usepackage{tabularx,wrapfig}
\usepackage{hyperref}

\usepackage{rotating}
\usepackage{color}           % para importar dibujos coloreados
\usepackage[all]{xy} % para diagramas

\usepackage[english]{babel}

\usepackage{graphicx}
\usepackage[utf8]{inputenc}
\usepackage{lscape}
\usepackage{color} 
\usepackage{url}

\theoremstyle{plain}% default

 \usepackage{latexsym} 
\usepackage{amsmath} % assumes amsmath package installed
\usepackage{amssymb}  % assumes amsmath package installed
\usepackage{rotating}

\theoremstyle{plain}% default

\usepackage{multicol}

\usepackage{pgfplots}
\pgfplotsset{compat=newest}

\usepackage{graphicx}
\usepackage{amsmath}
\usepackage{multirow}
\usepackage{tikz}
\usepackage{booktabs}

\usepackage{svg}
\usepackage{hyperref}

\newcommand{\orcid}[1]{\href{https://orcid.org/#1}{\includesvg[width=10pt]{figures/orcid}}}

\RequirePackage{stmaryrd}

\begin{document}

\journal{To decide}

\begin{frontmatter}

% Paper Title
\title{D3A-TS: Denoising-Driven Data Augmentation in Time Series}

% what is good, if it is simple, it is twice as good
% Authors List
% David: \orcid{0000-0002-4625-4554}
% Juan: \orcid{0000-0002-6622-271X}
% Joaquín: \orcid{0000-0003-0528-9459}
\author{%			
	David Solís-Martín$^{1,2}$ %\authorNumber{1,2} 
 %\orcid{0000-0002-4625-4554}, 
 Juan Galán-Páez$^{1,2}$
 %\orcid{0000-0002-6622-271X}, 
 and Joaquín Borrego-Díaz$^{2}$ 
 %\orcid{0000-0003-0528-9459}
}

% Author Affiliations

\address{% This is a tabular environment so each affiliation needs to be separated by "\\" or "\tabularnewline"
	$^1${Datrik Ingellicence, S.A. Seville, Spain}\\
 { %add emails
		{{\tt david.solis@datrik.com}},
		{{\tt juan.galan@datrik.com}}
		} \\ % emails input
	%\tabularnewline % skip one row for next affiliation	
	$^2${Department of Computer Science and Artificial Intelligence, Seville University, Seville, Spain}\\
 { % add emails
		{{\tt dsolis@us.es}},	{{\tt juangalan@us.es}},
		{{\tt jborrego@us.es}}
		} % emails input
}

% Create the title
\date{ }

% Abstract
\begin{abstract}
It has been demonstrated that the amount of data is crucial in data-driven machine learning methods. Data is always valuable, but in some tasks, it is almost like gold. This occurs in engineering areas where data is scarce or very expensive to obtain, such as predictive maintenance, where faults are rare. In this context, a mechanism to generate synthetic data can be very useful. While in fields such as Computer Vision or Natural Language Processing synthetic data generation has been extensively explored with promising results, in other domains such as time series it has received less attention. This work specifically focuses on studying and analyzing the use of different techniques for data augmentation in time series for classification and regression problems. The proposed approach involves the use of diffusion probabilistic models, which have recently achieved successful results in the field of Image Processing, for data augmentation in time series. Additionally, the use of meta-attributes to condition the data augmentation process is investigated. The results highlight the high utility of this methodology in creating synthetic data to train classification and regression models. To assess the results, six different datasets from diverse domains were employed, showcasing versatility in terms of input size and output types. Finally, an extensive ablation study is conducted to further support the obtained outcomes.

% \blue{Esto me ha dicho Claude cuando le he pedido que siga la estructura clásica, a partir del abstract anterior: 
% It has been shown that sufficient data quantity is crucial for data-driven machine learning approaches. In particular, in Engineering areas where data is scarce or very expensive to obtain, such as predictive maintenance, where faults are rare. A promising solution is the production of synthetic data. While synthetic data generation is well-explored in areas like computer vision, less attention has been given to time series data. This paper specifically focuses on studying and analyzing different techniques for time series data augmentation, for both classification and regression problems. The proposed approach employs recent diffusion probabilistic models, successfully used in image processing, to generate synthetic time series data. Additionally, the use of meta-attributes to condition the data augmentation is investigated.
% The results highlight the high utility of this methodology for creating synthetic data to train classification and regression machine learning models. Six datasets from diverse domains were used, exhibiting versatility. An extensive ablation study further supports the outcomes.
% Diffusion probabilistic models can effectively generate synthetic data for time series machine learning problems. The use of meta-attributes allows conditioning the data generation. This approach was validated across datasets of varying sizes and output types.}

\end{abstract}

% Revistas candidatas:
% - Neural Networks: https://www.sciencedirect.com/journal/neural-networks
% - Machine Learning: https://link.springer.com/journal/10994
% - Information Systems Frontiers: https://link.springer.com/journal/10796
% - Advanced Engineering Informatics: https://www.sciencedirect.com/journal/advanced-engineering-informatics

% - Information Sciences (D1?): https://www.sciencedirect.com/journal/information-sciences (no indica tiempo de aceptaciónp por el editor)
% - Applied Soft Computing: https://www.sciencedirect.com/journal/applied-soft-computing (no indica tiempo de aceptaciónp por el editor)

\end{frontmatter}

\section{Introduction}

%\red{Posible reorganización de los apartados. Lo he dejado comentado debajo de esta línea. Según hemos comentado, podemos dejarlo así y cambiar si los revisores dicen algo. Pero como ya lo tenía, lo dejo por aquí.}

% 3 Methods --> Background
% 3.1 Data Augmentation through denoising
% 3.2 Denoising Autoencoder
% 3.3 ¿Denoising? Diffusion models

% 4 Methodology?

% 3.4/4.1 Meta-attributes for conditioning denoising models
% 4.1/4.2 Datasets
% 4.2/4.3 Experiment settings
% ¿Algún subapartado?
% - calculo meta-atributos
% - Entrenamiento del denoising model, con y sin condicionamiento
% - construccion del modelo predictivo (clasificacion regresion)
% 4.4Bayesian test --> Result analysis/evaluacion/comparación de resultados

% 5 Main results

% 6 Ablation study

The advances in deep learning over the past few years have demonstrated that the amount of data used for training models is crucial. The more data available for training, the more generalized could be the model obtained; increasing the training data set size decreases the probability of overfitting. However, real-word data is limited \cite{wen2020time} and  the process of collecting and labeling data can be expensive \cite{iglesias2022data} \cite{liu2020efficient}. This limitation makes it impractical to obtain a large dataset, particularly in cases such as predictive maintenance where failure events occur infrequently. In such contexts, having a mechanism to expand the available data becomes crucial for improving the generalization of models.

One tool for increasing both the quantity and quality of data is known as data augmentation. Data Augmentation (DA) creates new data samples that remain representative with respect to the original, while introducing diversity into the training data set. The newly created samples must retain their labels in the case of classification tasks, or have targets as close as possible to the original target values in the case of regression problems.

In the case of Image Processing there are geometrical transformations, such as cropping, flipping, rotation, or translation, among others, that do not modify the category of the samples. In Natural Language Pocessing, a word can be replaced by a synonym \cite{mi2022improving}, or some words can be removed to create new sentences. However, these kind of transformations can not be applied directly in other scenarios. this is the case of time series data, where those transformations can not be used due to the particular nature of time series distributions. For example, a human can easily decide whether a transformed image or text still retains the original target. However, this is not that easy with time series data, since geometric transformations could potentially alter time-domain features \cite{wang2018data} without being detected.

%\blue{While a human can decide whether a transformed image or text still retains the original target, the same can not be applied to time series data, as geometric transformations could potentially alter time-domain features \cite{wang2018data} without being detected by a human. J: reescribir}

Noise addition has been widely used as a DA technique \cite{zur2009noise,moreno2018forward,xie2020unsupervised}. The method involves adding random noise to each training sample during training, what introduces variability to the training sample space,  hinding overfitting the training dataset. This method has been found to be equivalent to Tikhonov regularization \cite{bishop1995training}, making it intriguing to explore other related techniques for data enhancement. One approach previously studied consist in training a model to remove noise previously added to raw samples. This denoising model must learn the data distribution trough a training process to learn how to remove noise o reconstruct the raw sample. Since the random noise added can not be exactly predicted, new samples will be generated hoping to maintain the time domain features of the raw samples.

This work explores denoising models for data augmentation in time series, with a focus on both classification and regression tasks. Our contributions can be summarized as follows:

\begin{itemize}
    \item Introduction of a methodology D3A-TS for applying denoising models to data augmentation, validated across both classification and regression tasks in time series (section \ref{sec:da_denoising}).
    \item Investigation into the use of diffusion models for data augmentation in both classification and regression tasks, a field that lacks prior exploration (section \ref{sec:experiments}).
    \item Proposal and demonstration of the utilization of a set of meta-attributes to condition the denoising model, showcasing its efficacy in enhancing denoising models for data augmentation (section \ref{sec:meta-attributes}).
    \item Conducting a comprehensive ablation analysis of the proposed methods to validate our findings against raw data, noise augmentation, and autoencoders (section \ref{sec:ablation}).
\end{itemize}

\section{Previous work}

\begin{figure}[H]
\centering
\includegraphics[width=0.6\linewidth]{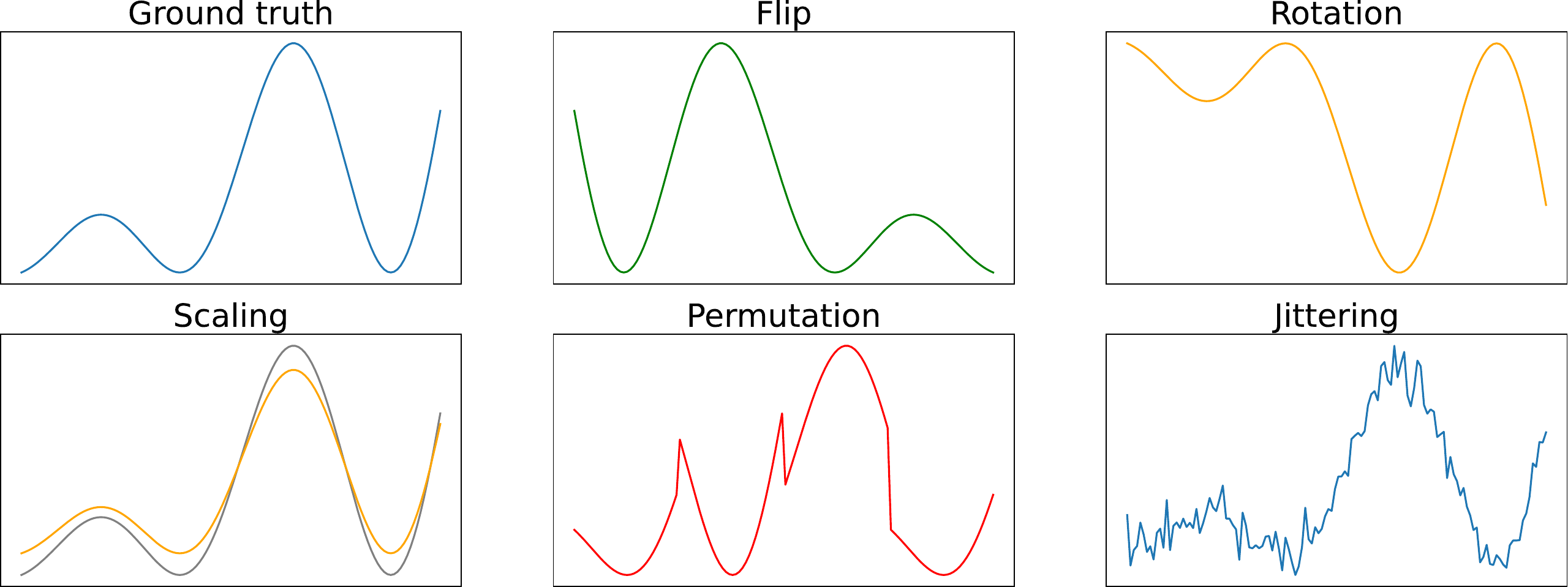}
\caption{A sample of basic data augmentation techniques.}
\label{fig:basic_da}
\end{figure}

Time series data augmentation methods need to possess the ability to both produce diverse samples and faithfully replicate the properties of real-world data \cite{wen2020time}. Fundamental DA techniques for time series are based on the manipulation of time series directly, for instance, deformation, modification, enlargement or shortening of the raw time series. Some of these algorithms are adaptations of data augmentation techniques from computer vision, to be applied in time series, and others are specifically designed for its use in this field. The main techniques are: flipping , jittering, scaling, rotation and permutation. These techniques are shown graphically in the figure \ref{fig:basic_da}.

\subsection{General Deep Learning models for DA}

Recent studies in generative Deep Learning (DL) models have made possible to create synthetic data of high quality. In this area three main architectures can be found: auto-enconders (AE), generative adversarial networks (GAN), and the recently successful diffusion probabilistic models (DPM). 

\subsubsection{Generative Adversarial Networks}

GAN architecture, introduced in 2014 by Ian Goodfellow \cite{goodfellow2014generative}, is composed by two networks: the generator and the discriminator. The goal of the generator network is to generate new samples from a random vector as similar as possible to the original ones, with the aim of misleading the discriminator. In the other hand, the discriminator has to predict whether the samples belong to the original samples or to the generated ones. During the training, the generator will become better on creating similar samples and the discriminator in detecting generated samples. In this way, the generator network, once it has been trained, allows extracting new samples from the learned data distribution. 
This type of architecture has been employed to augment data in various ways. In \cite{cabrera2019generative}, a GAN is used to tackle issues with extremely imbalanced datasets by generating samples for the less-represented class.
However, this architecture comes with an important drawback associated to the training phase \cite{jenni2019stabilizing,  roth2017stabilizing, arjovsky2017towards}. The basic GAN architecture does not require a reference raw sample to generate synthetic samples. Therefore, to assign a category to the generated sample, it is necessary to train a conditioned generator \cite{mogren2016c, esteban2017real, yoon2019time, lemley2017smart}. 

\subsubsection{Autoencoders}

The AE architecture was introduced by Hinton and the PDP group in 1986 \cite{rumelhart1985learning} to tackle the challenge of "backpropagation without a teacher", also known as unsupervised learning. This architecture consists of two networks: the encoder and the decoder. The encoder's primary goal is to obtain a low-dimensional latent representation of the input data, while the decoder's task is to reconstruct the original input using this low-dimensional representation. Due to the limited information in the low dimension representation of the input, one of the most popular applications of AEs is input denoising \cite{vincent2008extracting, gondara2016medical, jiang2017wind, cabrera2017automatic}. That is, random noise is added to the original sample and this noisy input is injected in the codificator. The output, given by the decodificator, will have the noise removed and therefore is expected to be very similar to the original sample. In addition, once the AE is trained, it is possible to use the decodificator to sample new data from the latent space. 

One problem with the AEs is that the latent space is not well structured, and therefore, the sampled data could have bad quality. There are other variants of AE that address this issue \cite{rudolph2019structuring}. Variational Auto-Encoders (VAE) \cite{kingma2013auto} mitigates this issue encoding the input using a Gaussian probability distribution. The latent vectors are generated from a vector of means and standard deviations, which define the latent space distribution. The imposition that the latent space must follow a gaussian distribution is achieved using the Kullback-Leibler Divergence as a regualization term in the loss function. 

VAEs have been used as a data augmentation tool in several works \cite{chadebec2021data, islam2021crash, nishizaki2017data}. Most of the applications of VAEs for data augmentation have been carried out conditioning the decoder to try ensuring the category of the sample generated. Only a few works have follow a denoising-based DA strategy using VAEs\cite{momeny2021learning}.

\subsubsection{Diffusion Probabilistic Models}

DPMs have recently garnered significant attention in the field of Computer Vision, showcasing remarkable achievements in image generation \cite{ho2020denoising}. These kind of model consists of two distinct stages: the \emph{forward diffusion} stage and the {reverse diffusion} stage. In the former, the input data undergoes a gradual transformation through the incremental introduction of Gaussian noise over multiple steps. In latter, a model is trained reconstruct the original input data by systematically removing the noise, step by step.

DPMs have been applied to various tasks in time series data, such as time series forecasting \cite{rasul2021autoregressive, li2022generative}, audio signal generation \cite{goel2022s, chen2020wavegrad}, and time series imputation \cite{tashiro2021csdi}. However, to the best of our knowledge, there is a notable absence of studies focused on the utilization of DPMs as a data augmentation technique for time series.

\section{Methods}

This section introduces the foundational instruments employed in the methodology for data augmentation in time series, developed in this work. The section unfolds through four subsections. It commences with an exploration of Data Augmentation through denoising (section \ref{sec:da_denoising}), leveraging techniques to feed the models with faithful samples that mirror real-world complexities. In the subsequent sections, Denoising Autoencoders (section \ref{sec:ae}) and Diffusion Model (section \ref{sec:dpm}), where the specific use of these models for both denoising and data augmentation approaches is introduced. In the last section (section \ref{sec:meta-attributes}), Meta-Attributes for Conditioning Denoising Models, the set of meta-attributes used to infuse and preserve sample characteristics, augmenting overall model performance, is defined.

\subsection{Data Augmentation through denoising} \label{sec:da_denoising}

Let $x^{(i)}$ be a time series sample from an input space $\mathcal{X} = \mathbb{R}^d$. A noisy sample, denoted as $\widetilde{x}_k$, is a sample that has noise added through some mapping process:

$$\widetilde{x_k} = \mathcal{M}(x) = x + \xi_k$$

Where $\xi_k$ is sampled from some noise distribution and $k$ denotes a sample extraction step. Usually $\xi_k$ is sampled from a Gaussian distribution with mean of 0 and standard deviation of 1;  $\xi_k \sim \mathcal{N}(0,I)$. Note that $\xi_k$ can be decomposed in the summation of lower-rate noises $\xi _{k} = \sum_{i=0}^T \epsilon_{k,i}$. In this way, a noisy sample can be seen as a process of progressive noise insertion $\widetilde{x}_{k,1}, \widetilde{x}_{k,2}, .., \widetilde{x}_{k,t}, .., \widetilde{x}_{k,T}$ where $\widetilde{x}_{k,t} = \sum_{i=0}^t \epsilon_{k,i}$.

A denoising model $\mathcal{H}$ can be trained to learn to remove noise introduced, denoted as $\xi_{k}$. That is, essentially, reconstructing the sample $x$ from the noisy sample $\widetilde{x}_k$, where $x \approx \mathcal{H}(\widetilde{x}_k) = \hat{x}_k$. Where $\hat{x}_k$ represents the reconstructed sample by the model $\mathcal{H}$ and is an approximation of the source sample $x$. Another approach involves training the model to remove only part of the noise introduced, i.e., $$\widetilde{x}_{k, t-1} \approx \mathcal{H}(\hat{x}_{k, t}, t) = \hat{x}_{k, t-1}$$ 

When the denoising model $\mathcal{H}$ has learned the data distribution and the noise added is sufficiently small, the category of the sample $\hat{x}$ should be likely retained.

Another way to enforce the preservation of the category is by conditioning the denoising model on the category of the sample $x$, expressed as $\mathcal{H}(\cdot, y) = \hat{x}{\cdot}$, where $y$ denotes the category of the sample $x$. This approach can be generalized by conditioning the denoising process on a set of meta-attributes extracted from the sample: $\mathcal{H}(\cdot, a) = \hat{x}{\cdot}$, where $a = \mathcal{A}(x)$, with $\mathcal{A}$ representing the process of meta-attribute extraction and $a$ being the meta-attributes vector extracted from $x$.

%If the mapping process $\mathcal{M}$ is stochastic, multiple samples can be generated: 

%$$\{ \mathcal{M}(x, 1), \mathcal{M}(x, 2), .., \mathcal{M}(x, k), ..,  \mathcal{M}(x, n) \} =%
%$$\{\widetilde{x}_1, \widetilde{x}_2, .., \widetilde{x}_k, .., \widetilde{x}_n \}$$ 

If the mapping process $\mathcal{M}$ is stochastic, multiple noisy samples can be generated. This set of noisy samples can then be denoised using the model $\mathcal{H}$ to produce a set of new samples, which are likely to belong to the same category than the source sample $x$:

$$\{\mathcal{H}(\mathcal{M}(x, \cdot)), \mathcal{H}(\mathcal{M}(x, \cdot)), .., \mathcal{H}(\mathcal{M}(x, \cdot)), .., \mathcal{H}(\mathcal{M}(x, \cdot)) \} = \{\hat{x}_{1, \cdot}, \hat{x}_ {2, \cdot}, .., \hat{x}_{k, \cdot}, .., \hat{x}_{n, \cdot}\}$$

Therefore, in this way it is possible to extend the data set generating $n$ new samples from each sample $x$. For simplicity, in the rest of the paper the $k$ parameter will be ommited.

\subsection{Denoising Autoencoder} \label{sec:ae}

A Denoising Autoencoder (DAE) is an autoencoder designed to reconstruct the original sample $x^{(i)}$ from a noisy observation $\widetilde{x}^{(i)}$. During the training phase, the model's parameters need to be optimized by minimizing the average reconstruction error, which is typically represented by the mean squared error, as follows:

\begin{equation}
\underset{\phi,\theta}{\text{arg min}} \frac{1}{M} \sum_{i=1}^{M} \mathcal{L}\left(x^{(i)} - f_\phi \cdot g_\theta\left(\widetilde{x}^{(i)} \right)\right)
\end{equation}

In this equation, $\mathcal{L}$ denotes the reconstruction error, $g$ and $f$ represent the encoder and decoder networks, respectively, and $\phi$ and $\theta$ denote the parameters of the encoder and decoder networks that need to be learned.

Once the DAE is trained, multiple observations based on $x^{(i)}$ can be obtained using the DAE:

\begin{equation}
\hat{x}^{(i)} = f_\phi \cdot g_\theta( \widetilde{x}) 
\end{equation}

Noted that:

$$\mathcal{H} = f \cdot g$$

In this work is considered that the demonising process with the autoencoder can be applied multiples times over the same sample. This is denoted as:

\begin{equation}
\begin{split}
& \hat{x}_{t-1}^{(i)} = f_\phi \cdot g_\theta( \hat{x}_{t}^{(i)});  T \geq  t \geq  1
\end{split}    
\end{equation}

where $t$ denotes the remaining denoising steps.

\subsection{Diffusion models} \label{sec:dpm}

In a broader sense, diffusion models adhere to the concept of systematically eliminating noise over a series of steps. These models are trained to predict the noise introduced at each step of the process. Consequently, they can be employed to progressively remove the noise, step by step. The procedure of introducing noise is referred to as the forward diffusion process, and its corresponding posterior distribution is denoted as $q$:

\begin{equation}
\begin{split}
 q(x_{1:T}|x) &= q(x_0)q(x_1|x_0)...q(x_{T}|x_{T-1}) \\
&= q(x_0)\mathcal{N}(x_1|\sqrt{1 - \beta_1}x_0, \beta_1 I)...\mathcal{N}(x_{T}|\sqrt{1 - \beta_{T}}x_{T-1}, \beta_{T} I) \\
&= q(x_0) \prod_{t=1}^T  \mathcal{N}(x_{t}|\sqrt{1 - \beta_{t}}x_{t-1}, \beta_{t} I)
\end{split}
\end{equation}
where $I$ is the identity matrix and $\beta_1, \beta_2, ..., \beta_T$ represents an increasing variance schedule with $\beta_t \in (0,1)$ controlling the level of noise added at each step. $\mathcal{N}(x|\mu, \sigma)$ denotes the probability density at $x$. It is worth noting that as $T$ approaches infinity, it resembles an isotropic Gaussian distribution.

By using the parametrization trick, it is possible to sample $x_t$ at any arbitrary time step in a closed form, avoiding the need to execute the noise addition process over T steps. Let $\alpha_t = 1 - \beta_t$, and $\sqrt{\overline{\alpha}_t} = \prod_{i=1}^t \alpha_i$:

\begin{equation}
    q(x_t|x_0) = \mathcal{N}(x_{t}\sqrt{\overline{\alpha}_t}x_{t-1}, \overline{\alpha}_t I)
\end{equation}

In order to sample new data, it is necessary to learn the reverse distribution $q(x_{t-1}|x_t)$. However, obtaining the closed form of the former posterior distribution is intractable, as it would require the computation of the involved data distribution. This issue can be solved by training a parameterized model $p_{\phi}$ to approximate $q(x_{t-1}|x_t)$. The model $p_{\phi}$ can be set to be Gaussian:

\begin{equation}
    p_{\phi}(x_{t-1}|x_t) = \mathcal{N}(x_{t-1}|\mu_{\phi}(x_t, t), \sigma_{\phi}(x_t, t) I)
\end{equation}
where $\phi$ represents the parameters of the model to be learned. Therefore, the model $p_{\phi}$ will be trained to predict  both the mean $\mu_{\phi}$ and variance $\sigma_{\phi}$ for time step $t$. 

The training process requires minimizing the negative log-likelihood of the training data. In \cite{ho2020denoising}, authors introduce a loss function that performs better than the original Evidence Lower Bound (ELBO), denoted as:

\begin{equation}
    \underset{\phi}{\text{arg min}} \frac{1}{M} \sum_{i=1}^{M} \mathcal{L} (\epsilon - \epsilon_{\phi} (\sqrt{\overline{\alpha}}_t x_0 + \sqrt{1 - \overline{\alpha}}_t\epsilon, t) )
\end{equation}

%\begin{equation}
%    \mathcal{L} = \mathbb{E}_{x_0, t, \epsilon} \left [ \parallel \epsilon - \epsilon_{\phi} (\sqrt{\overline{\alpha}}_t x_0 + \sqrt{1 - \overline{\alpha}}_t\epsilon, t) \parallel^2   \right ]
%\end{equation}

where $\epsilon \sim \mathcal{N}(0,I)$. The previous equation aims to minimize the noise $\epsilon$, added to the sample $x_{t-1}$, and the error predicted by $\epsilon_{\phi}$, given a noisy sample, represented as $\sqrt{\overline{\alpha}}_tx_0 + \sqrt{1 - \overline{\alpha}}_t\epsilon$ and the time step $t$. This can be denoted as:

% validar esto:
\begin{equation}
    p_{\phi}(x_{t-1}|x_t) = \frac{\sqrt{\overline{\alpha}_{t}} x_0 + \sqrt{1 - \overline{\alpha}_{t}}\epsilon - \sqrt{1 - \overline{\alpha}_{t}} \epsilon_{\phi} (\sqrt{\overline{\alpha}}_t x_0 + \sqrt{1 - \overline{\alpha}}_t\epsilon, t) }{\sqrt{\overline{\alpha}_{t}}} 
\end{equation}

% https://www.assemblyai.com/blog/diffusion-models-for-machine-learning-introduction/
% https://theaisummer.com/diffusion-models/

% the training loss can be expressed as a regression problem when we assume that the denoising process is given by a learned diagonal Gaussian.

\subsection{Meta-attributes for conditioning denoising models} \label{sec:meta-attributes}

As mentioned earlier, a denoising model $\mathcal{H}$ can be conditioned on a set of meta-attributes $a$. In this work, the hypothesis under study is that this conditioning can enhance the denoising model's effectiveness in the data augmentation process by assisting in preserving the source category of the sample. 

For each time series, 15 meta-attributes were extracted. Note that, since the work is focused on time series data, the meta-attributes has been specifically selected to work with time series. These are described as follows:

\begin{itemize}
    \item \textbf{Stability}. Squared Pearson correlation coefficient. 
    \item \textbf{Periodicity}. The mean Pearson correlation between four equally sized  consecutive segments of the sample.  
    \item \textbf{Oscillation}.  The absolute of the ratio between the standard deviation and the mean of the sample. 
    \item \textbf{Complexity}. The sum of the absolute values of the Fourier transformation magnitudes. 
    \item \textbf{Noise}. The ratio between the mean and the standard deviation of the sample. 
    \item \textbf{Entropy}. The Shannon's entropy of the sample's probability distribution. 
    \item \textbf{Variability}. The ratio between the standard deviation and the mean of the sample. 
    \item \textbf{Standard deviation}. The standard deviation of the sample  
    \item \textbf{Peculiarity}. The sample kurtosis (calculated with the adjusted Fisher-Pearson standardized moment coefficient G2  \cite{christ2018time}
    \item \textbf{Dynamic range}.  The absolute value of the difference between the maximum and minimum values of the sample. 
    \item \textbf{Symmetry}. The sample skewness (calculated with the adjusted Fisher-Pearson standardized moment coefficient G1). \cite{christ2018time}
    \item \textbf{Peaks}. The number of peaks that occur at enough width scales and with sufficiently high Signal-to-Noise-Ratio (SNR)  \cite{christ2018time}. This feature is computed from the smoothed sample using the ricker wavelet for widths ranging from 1 to 10. 
    \item \textbf{Slope}. The slope of the linear regression of the sample 
    \item \textbf{Min value}. The minimum value of the sample 
    \item \textbf{Max value}. The maximum value of the sample         
\end{itemize}

% Nota: incluir ejemplos de la señal y con los metaatributos

\section{Experiments} \label{sec:experiments}

This section outlines the methodology employed for conducting experiments. Initially, a brief description of the datasets used is provided. Subsequently, the approach adopted to conduct the experiments is explained.

\subsection{Datasets}

Six different datasets have been considered to carry out the experiments on different type of problmes. Two of these datasets belong to regression tasks, two to binary classification, and the remaining two to multiclass classification. Below, each dataset will be briefly described.

\subsubsection{N-CMAPSS}

The N-CMAPSS dataset \cite{ref-arias} have been built using the Commercial Modular Aero-Propulsion System Simulation (CMAPSS), a modeling software developed at NASA. N-CMAPSS provides full flights trajectories starting with a healthy condition until a failure occurs. This dataset is gaining popularity in the context of prognosis and health management (PHM), where works such as \cite{solis2021stacked, solis4596645deep, cohen2023shapley}, among others, have proposed different solutions to address the problem posed by this dataset. 

The dataset presents seven failure modes associated with either flow degradation or subcomponent efficiency. The flights are categorized into three classes based on their duration: Class 1 includes flights lasting 1 to 3 hours, Class 2 covers flights ranging from 3 to 5 hours, and Class 3 comprises flights lasting over 5 hours. 
%Furthermore, each flight is subdivided into cycles, encompassing the stages of climb, cruise, and descent operations. 
The problem to be addressed with this dataset is to predict the remaining useful life (RUL) of the system until a failure occurs. It is assumed that RUL will decrease linearly from the maximum value (total cycles of the experiment) to 0. The dataset provides 20 input time series variables, therefore this is a multivariate regression task. 

\subsubsection{PRONOSTIA}

PRONOSTIA dataset\cite{ref-pronostia}, collected by FEMTO-ST, a French research institute, using the so called PRONOSTIA platform. The bearings mounted on this platform were subjected to gradual degradation over time. To expedite this degradation process, three distinct operating conditions were considered, which respectively involved rotating speeds of 1800 rpm with a 4000 N payload weight, 1650 rpm with 4200 N, and 1500 rpm with 5000 N.

Two sensors, placed along the x-axis and y-axis of the bearing, were employed for data acquisition. Data were recorded every 10 seconds during 0.1 seconds, with a frequency of 25.6 kHz, therefore, resulting in each time series comprising 2560 data points. The experiment concluded when the vibration amplitude exceeded the threshold of 20 g. Similar to the previous dataset, it is assumed that the remaining useful life (RUL) ofthe system decreases linearly from the maximum value (total time of the experiment) to 0. However, for this study, the RUL is normalized to be between 0 and 1. 

\subsubsection{ECG5000}

The ECG5000 dataset, sourced from the University of California, Riverside, serves as a prominent benchmark for time series classification \cite{goldberger2000physiobank}. It stands as one of the most widely used public datasets for time series classification. The dataset comprises a 20-hour long ECG (Electrocardiogram) data stream, and encompasses 5 distinct categories. Each data instance is a sequence with a length of 140 data points. 

The dataset consists of 500 samples for training and 4500 for testing.

\subsubsection{Human Activity}

% Human Activity
The Human Activity (ha dataset (HR) \cite{anguita2013public} is focused on Human Activity Recognition (HAR). The experiments, carried out to build the dataset, involved a group of 30 volunteers, each performing six activities (walking, walking upstairs, walking downstairs, sitting, standing and laying) while wearing a smartphone (Samsung Galaxy S II) at the waist. The accelerometer signal from the smartphone was captured, preprocessed to eliminate noise, and divided into 128-point time series, each associated with a specific action. The training set comprises 7352 samples, while the test set consists of 2947 samples.

\subsubsection{Wine}

The Wine dataset \cite{timeseriesclassification} consists of spectrographs from two types of wine: cabernet sauvignon and shiraz. These signals were obtained using FTIR spectroscopy with attenuated total reflectance (ATR) sampling. 

The dataset contains 57 samples for training and 54 for testing, with each sample comprising 234 data points. 

\subsubsection{Shares}

Finally, the Shares dataset \cite{timeseriesclassification} is designed for predicting whether a share price will experience a notable increase after the quarterly announcement of earnings, based on the price movement in the preceding 60 days. Each data point represents the percentage change in the close price with respect to the day before over a 60-day period. The target class is assigned a value of 1 if the company stock increased by more than 5 percent after the report release, and 0 otherwise. The dataset comprises 1931 samples, with 1326 instances classified as class 0 and 605 as class 1.

\begin{figure}[t]
    \centering
    \includegraphics[width=0.6\textwidth]{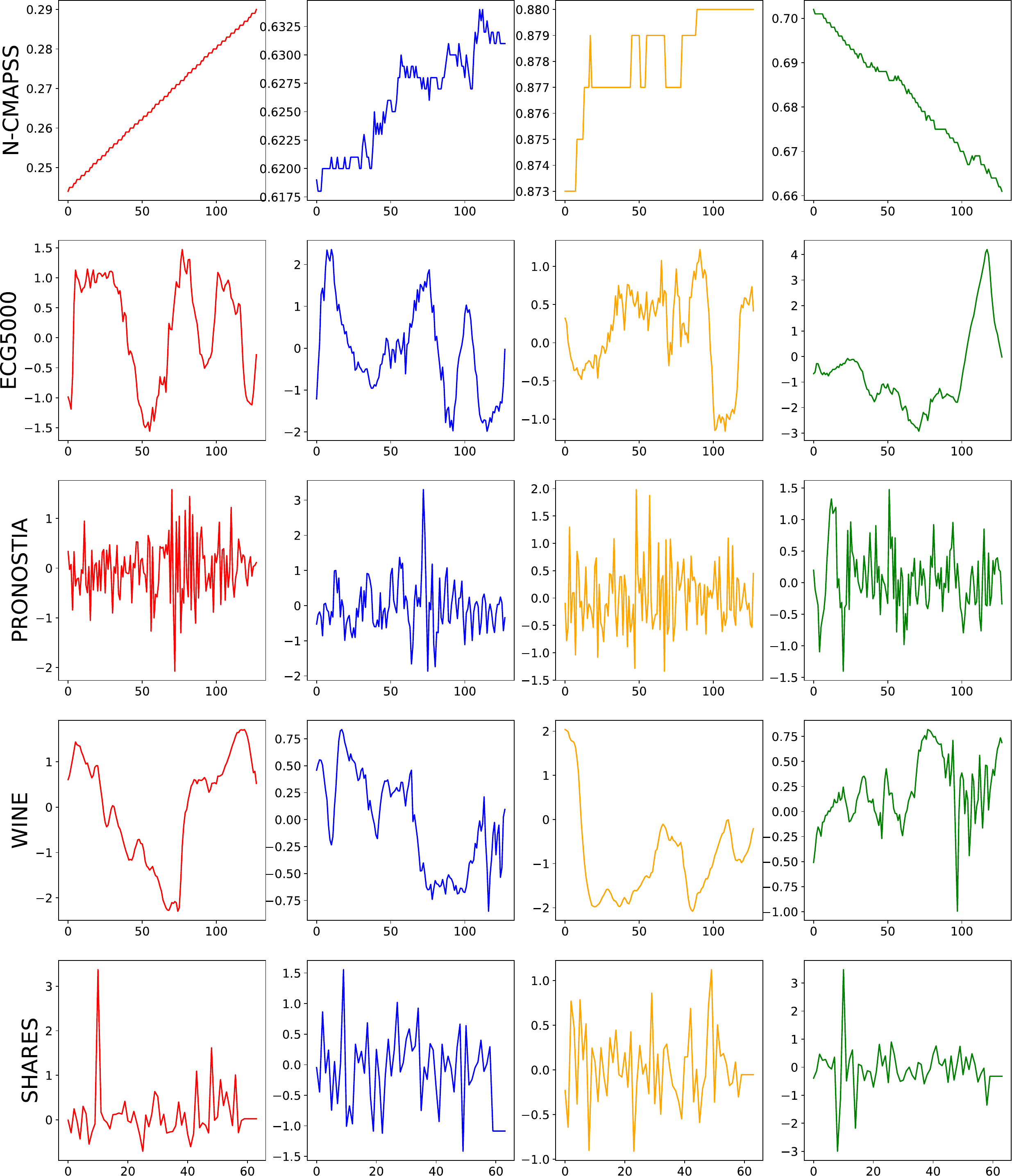}
    \caption{Samples from the six datasets used in this work. It can be observed that there exists a lot of diversity in the shapes of the samples.}
    \label{fig:dataset_signals}
\end{figure}

Table \ref{table:datasets} summarizes the main features of the selected datasets. These datasets have been sourced from different areas, involve diverse tasks, and vary in dimensions, to cover a wide variety of experiments. Figure \ref{fig:dataset_signals} showcase a few samples from each dataset, highlighting the diversity of shapes explored in this paper.

\begin{table}[]{\footnotesize
\begin{tabular}{lllrrcrr}
\toprule
 \multicolumn{1}{c}{\textbf{Dataset}}       &  \multicolumn{1}{c}{\textbf{Task}}  &  \multicolumn{1}{c}{\textbf{Area}}    & \multicolumn{1}{c}{\textbf{I.D.}} & \multicolumn{1}{c}{\textbf{Length}} & \multicolumn{1}{c}{\textbf{O.R.}} & \multicolumn{1}{c}{\textbf{Train / Test size}}  \\
\midrule
\textbf{N-CMAPSS}       & R     & I       & 20                                           & \textgreater 3e5                    & {[}0-100{]}                                 & 195e5                           / 840e5                     \\
\textbf{PRONOSTIA}      & R     & I       & 2                                            & 2560                                & {[}0-1{]}                                 & 10240                                   / 7680                          \\
\textbf{SHARES}         & B         & F        & 1                                            & 60                                  & \{0,1\}                                   & 965                                     / 965                           \\
\textbf{WINE}           & B         & P          & 1                                            & 234                                 & \{0,1\}                                   & 57                                      / 54                            \\
\textbf{ECG5000}        & M (5) & H           & 1                                            & 140                                 & \{0,1,2,3,4\}                             & 500                                     / 4500                          \\
\textbf{HR} & M (6) & CS & 1                                            & 128                                 & \{0,1,2,3,4,5\}                           & 7352                                    / 2947                         \\
\bottomrule
\end{tabular}}
\caption{Summary of the main features for the six datasets used in this work. I.D.: Input Dimension, O.R.: Output Range, R: Regression, B: Binary, M (x): Multi class with x categories, I: Industry, F: Financial, P: Physics, H: Health, CS: Computer Science.}
\label{table:datasets}
\end{table}

\subsection{Experiment settings}

Before executing any experiment, it is necessary to generate the meta-attributes $a^{(i)}$ associated with each sample $x^{(i)}$. In this study, the samples have been truncated to have 128 data points (64 data points in the case of the shares dataset). For each segment of 32 data points, 15 meta-attributes are extracted. As a result, the vector $a^{(i)}$  will have a size of 60 elements (30 in the case of the shares dataset). It's important to note that this process is computationally intensive.

To address this computational complexity, the meta-atributes vectors $a^{(i)}$  are approximated using a single fully connected neural network $\mathcal{A}_{\psi}$. The network $\mathcal{A}_{\psi}$ comprises three hidden layers with 32, 64, and 128 neurons, where the hyperbolic tangent (tanh) function is utilized as the activation function.

\begin{figure}[t]
    \centering
    \includegraphics[width=0.8\textwidth]{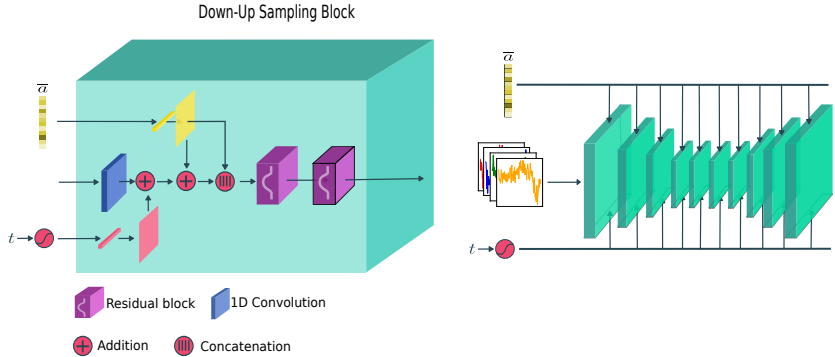}
    \caption{Left: the upsampling and downsampling block used to create the denoising model architectures for both autoencoders and diffusion models. Right: the denoising model architecture composed of four downsampling steps, two additional blocks, and four upsampling steps.}
    \label{fig:arch_net}
\end{figure}

The model $\mathcal{A}_{\psi}$ is used while training the denoising model $\mathcal{H}_{\phi}$ to learn to remove the noise from a noisy sample. The model $\mathcal{H}_{\phi}$ is composed of 4 downsampling blocks and 4 upsampling blocks. After downsampling, two additional blocks are applied (see Figure \ref{fig:arch_net}). Each block applies one-dimensional operations to the input. Subsequently, the time step embedding and the attribute embedding are integrated with the feature map using sum and concatenation operations. Then, two residual blocks are applied. These embeddings are created using a fully connected layer and the corresponding reshaping to match the dimensions of the former one-dimensional convolutional layer.

\begin{figure}[t]
    \centering
    \includegraphics[width=0.8\textwidth]{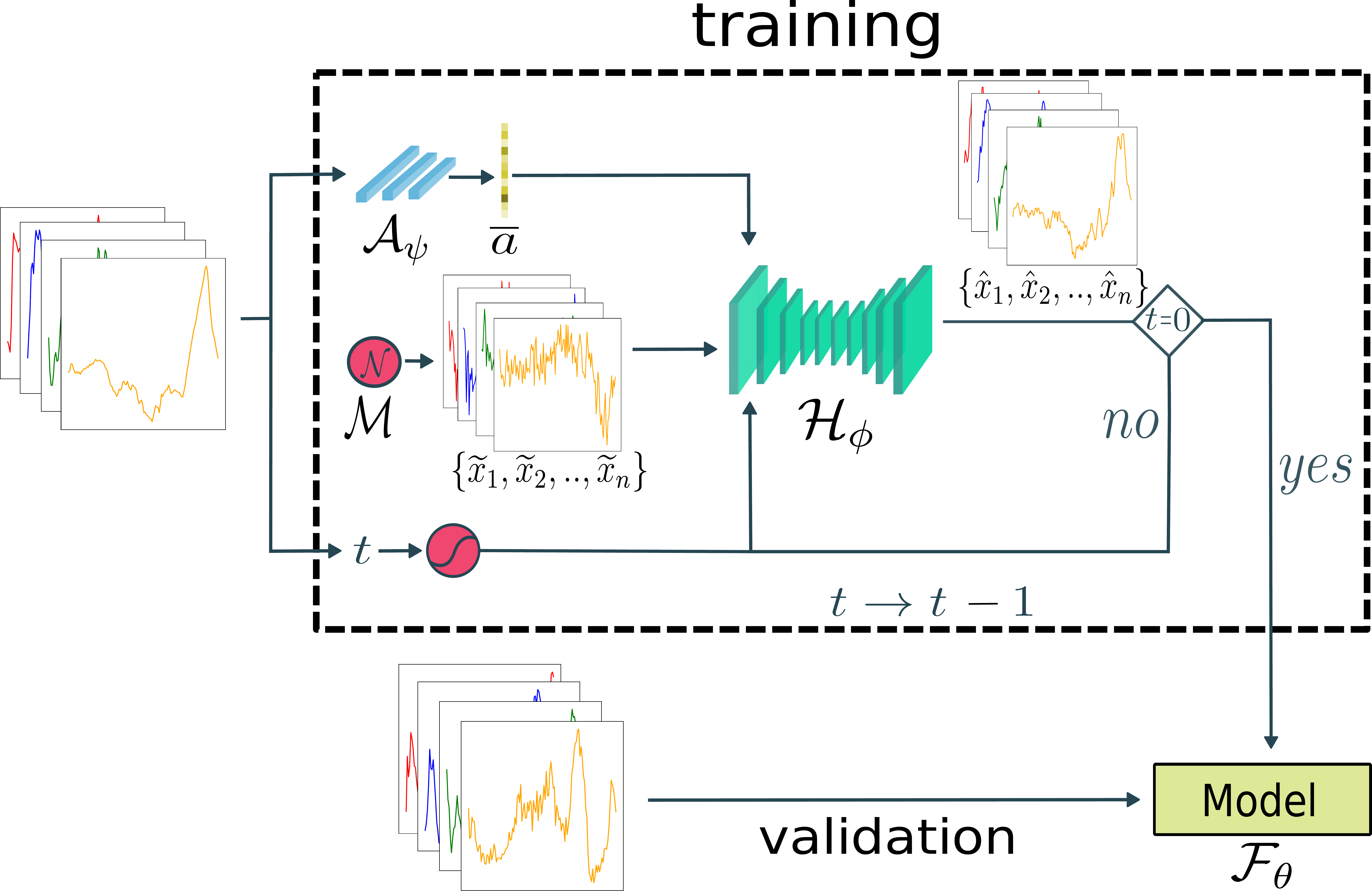}
    \caption{The graph illustrates the training process. $\mathcal{A}_{\psi}$ represents the network used to predict the meta-attributes vector $a$ from the training raw data. $\mathcal{M}$ denotes the process that introduces normal noise to the raw samples, while $\mathcal{H}_{\phi}$ represents the denoising network responsible for generating the synthetic samples. The model $\mathcal{F}$ is then trained using these synthetic samples and validated against the raw samples from the test set.}
    \label{fig:training}
\end{figure}

Finally, the model $\mathcal{H}_{\phi}$ is used to augment data during the training of the model to solve the classification or regression task. It is worth noting that this final model is exclusively trained using synthetic data. These final models are trained using 1, 2, and 3 denoising steps, along with various noise rates introduced to the samples before applying denoising. Both the number of denoising steps and the noise rate are considered hyper-parameters of the training process. %The model $\mathcal{H}_{\phi}$ applied recursively $n$ steps is denoted as $\mathcal{H}_{\phi}^n$. % Jaoquín dice que no es necesario
The stopping criterion for the training procedure is early stopping, which halts the training of a model once its performance on a validation set has not improved after a predefined number of training steps. This process is summarised in the algorithm \ref{algo:training} and depicted in figure \ref{fig:training}.

\begin{algorithm}[H]
\caption{Training algorithm}\label{alg:cap}
\begin{algorithmic}
\State Set $n$ as the number of denoising steps.
\State Set $r$ as the noise ratio.
\State $X$ is a set of samples.
\State $Y$ is the set of target values $y^{(i)}$ for each $x^{(i)} \in X$
\State $A$ is a set of meta-attributes extracted from each sample $x^{(i)} \in X$
\State $\mathcal{M}(x, r)$ is a mapping process that introduces noise with the form $r\mathcal{N}(0, 1)$  
\State Train a model $\mathcal{A}_{\psi}$ to predict $a^{(i)} \in A$ from $x^{(i)} \in X$
\State Train a denoising model $\mathcal{H}_{\phi}(x_{t-1}^{(i)},\mathcal{A}_{\psi}(x^{(i)}))$ to predict $x_{t}^{(i)}$
\State Train a model $\mathcal{F}_{\theta}(\mathcal{H}_{\phi}(\mathcal{M}(x^{(i)}, r), \mathcal{A}_{\psi}(x^{(i)}))$ to predict $y^{(i)} \in Y$
\end{algorithmic}
\label{algo:training}
\end{algorithm}

In addition to the final model trained with denoised (synthetic) data, final models using the raw and noisy samples ,respectively, have also been trained to be compared with such a model. Different experiments were carried out in which, DPM (Denoising Process Models) and AE (Autoencoder) models are used as denoising models.

Two different network architectures were tested with each dataset. The first one is a multi-scale convolutional neural network \cite{cui2016multi} with two multi-scale blocks and two additional convolutional blocks, each composed of two stacked convolutional layers and max-pooling operations. Following the convolutional layers, two fully connected layers are added before the output layer. The second architecture is a recurrent network composed of two stacked bidirectional LSTM layers and two fully connected layers. Mean squared error was used as the loss function for regression, and softmax loss was used for classification.

The complete set of experiment settings along with the results obtained in each one and the source code used to execute all these experiments can be checked in the GitHub repository \url{https://github.com/DatrikIntelligence/D3A-TS-Denoising-Driven-Data-Augmentation-in-Time-Series}.

\section{Bayesian test}

Due to the well-documented issues with the null hypothesis significance test \cite{benavoli2017time}, this work adopts a Bayesian approach to confront the results obtained. Specifically, the Bayesian signed-rank test \cite{benavoli2014bayesian} is used, which serves as a Bayesian adaptation of the Wilcoxon signed rank test. This test takes as input the paired results of two models, A and B, and provides three probabilities: the likelihood that model A is better than B, the likelihood that model B is better than A, and the probability that there are no significant differences (rope).

The ratio between model A and model B is adopted as the paired input, computed as:
\begin{equation}
    \mathcal{R}(a,b) =  \frac{a - b}{0.5(a+b)}
\end{equation}
Here, $a$ and $b$ represent the losses obtained by models A and B, respectively. $R$ will take a negative value when model A outperforms model B (the loss error of A is lower than the loss error of B), and a positive value in the opposite case. It is considered that a value of $R \in [-0.05,0.05]$ means that there are not significant differences (rope).

\section{Main results}

A considerable number of experiments have been executed, involving six datasets, 2 final networks, 3 data augmentation techniques, and around 10 different noise rates for the 2 denoising models, each with 3 repetitions per trained model, which makes a total of 1800 experiments. The results highlight that using the diffusion model for data augmentation in time series problems, for both classification and regression, is entirely plausible. Additionally, the results demonstrate that the use of meta-attributes improves the behavior of the denoising models by preserving the sample category, as hypothesized in Section \ref{sec:meta-attributes}.

\begin{figure}[t]
    \centering
    \includegraphics[width=0.8\textwidth]{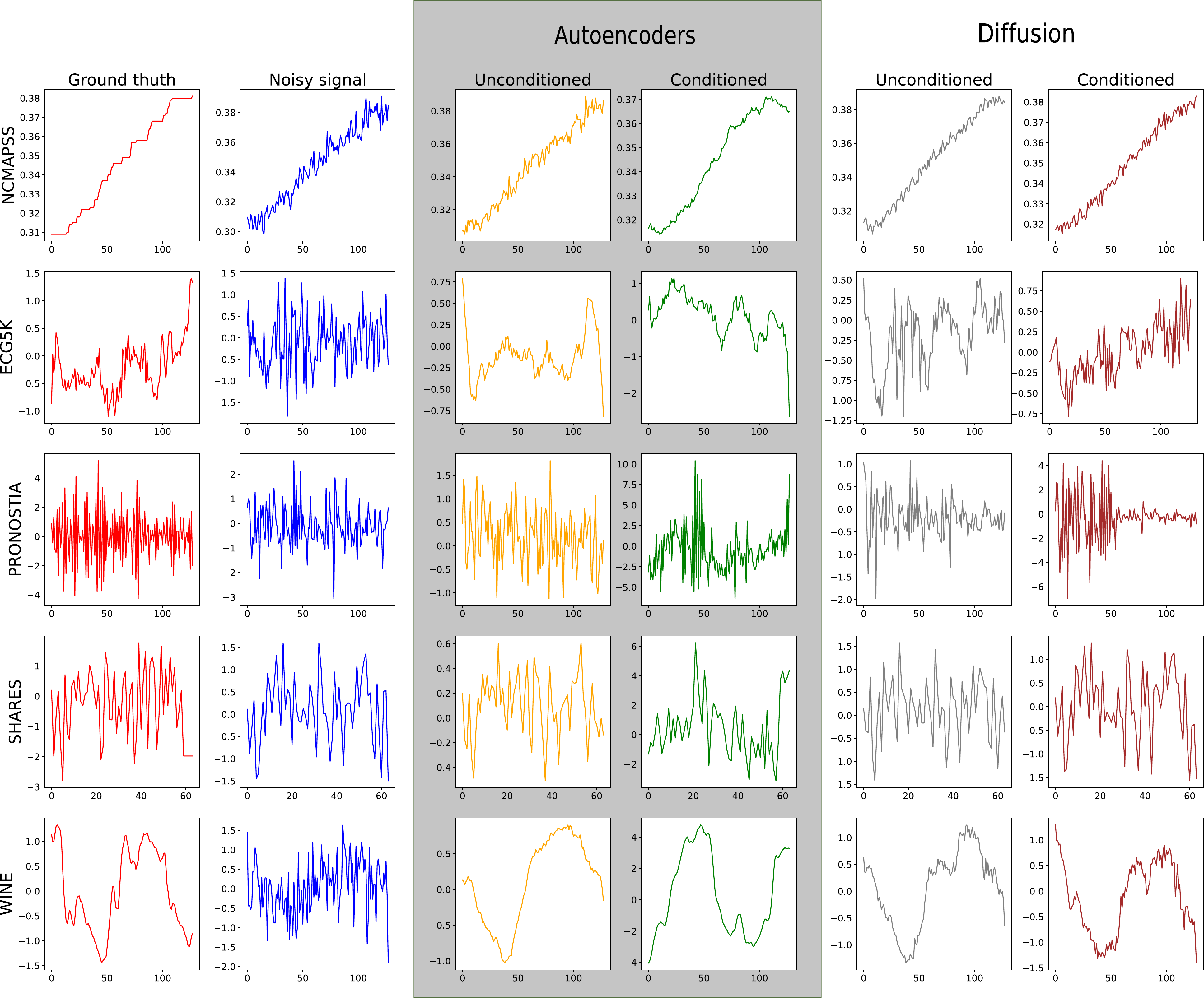}
    \caption{Samples from the six datasets used in this work, along with examples showing the noisy samples after noise addition and the reconstructions made by the autoencoders and diffusion models, both with unconditional and conditional denoising.}
    \label{fig:denoise_samples}
\end{figure}

Figure \ref{fig:denoise_samples} displays examples of raw samples, along with their noisy version and the results obtained after applying denoising with autoencoders and diffusion models. The autoencoders and diffusion models have been trained both in an unconditional and conditional way.

It is important to emphasize that the primary objective  of this work is not to achieve the model with the lowest loss for each datasets. The focus is primarily on assessing whether the denoising process can lead to better generalization of a model, irrespective of the architecture, hyperparameters, and the data involved. Consequently, the hyperparameters of the regression or classification model have not been extensively optimized. For each dataset, different experiments have been carried out to evaluate the proposed methods, that is, denoising data augmentation based on AE and DPM, with and without meta-attributes conditioning. Additionally, models have been trained with noise-based data augmentation. For each of these experiments, another has been performed using raw data, to assess whether denoising data augmentation based models improve performance. Each experiment has been repeated three times to compute the mean loss and compare the results among the different approaches. In the case of data augmentation methods (noise, AE, and DPM), different noise rates $\beta_t$ have been tested. Finally, tor denoising methods, 1, 2, and 3 denoising steps $t$ have been evaluated.

%\red{TO-DO DAVID: Añadir comentario aqui indicando que se realizará un experimento por cada valor de beta y T.}

\begin{figure}[t]
    \centering
    \includegraphics[width=1\textwidth]{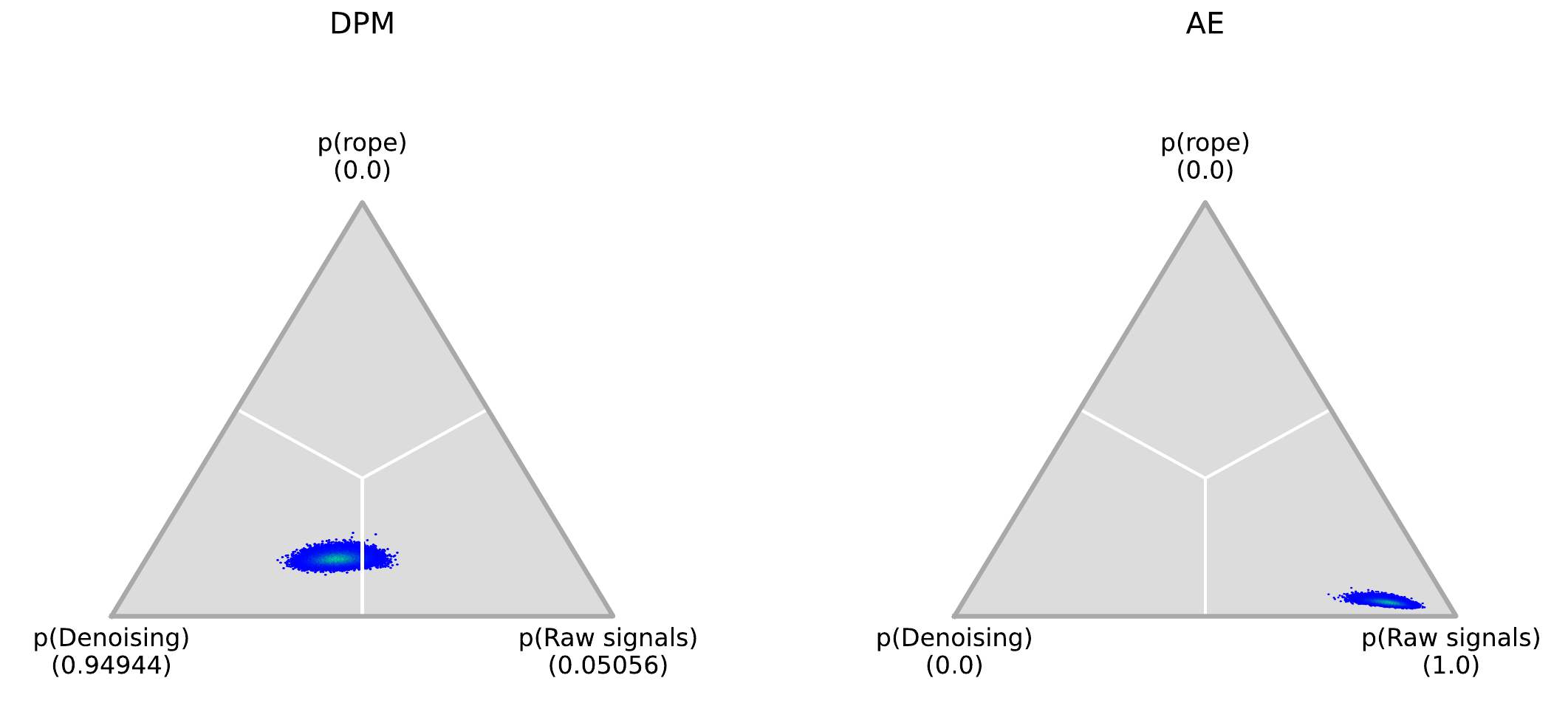}
    \caption{Comparative performance of a network trained with denoising data augmentation and the same network trained with raw data.}
    \label{fig:denoise_vs_raw}
\end{figure}

Table \ref{table:denoise_vs_raw} presents the results achieved using denoising conditioned data augmentation, which are better than the average performance of the models trained using raw data. As observed, in all instances, it is possible to identify hyperparameters that surpass the models trained solely on raw data.  Additionally, in most cases, a specific set of hyperparameters demonstrates that the DPM data augmentation outperforms the AE data augmentation. Figure \ref{fig:denoise_vs_raw} shows the results of the Bayesian signed-rank test comparing DPM and AE conditioned data augmentation with the models trained using raw data. The Bayesian signed-rank test confirms that finding good hyperparameters in DPM has a high probability, while using AE, the probability is very low.

\begin{table}[]{\footnotesize
\begin{tabular}{llrrr}

\toprule
       Dataset &    Net &                  Raw &                      AE &                     DPM \\
\midrule
     ecg5k & rnn &   0.5669 ±   0.1651  &   0.6932 ±   0.4601 [1] &   \textbf{0.3235} ±   0.0085 [1] \\
         ecg5k &  cnn &   0.6631 ±   0.2387  &   0.5204 ±   0.0325 [3] &   \textbf{0.4272} ±   0.0205 [1] \\
HA & rnn &   1.2370 ±   0.2486  &   1.1023 ±   0.0016 [1] &   \textbf{1.0897} ±   0.0011 [2] \\

HA &  cnn &   1.2821 ±   0.1238  &   \textbf{1.2461} ±   0.0221 [2] &   1.2758 ±   0.0120 [2] \\
       ncmapss & rnn & 252.9270 ±  13.9745  & \textbf{242.1542} ±   0.0000 [2] & 246.7824 ±  16.9870 [2] \\
       ncmapss &  cnn & 459.5337 ± 165.5178  & 324.1614 ±  16.9442 [2] & \textbf{266.0854} ±  13.5774 [1] \\
     pronostia & rnn &   0.0720 ±   0.0063  &   0.0700 ±   0.0013 [1] &   \textbf{0.0662} ±   0.0007 [1] \\
     pronostia &  cnn &   0.0614 ±   0.0044  &   \textbf{0.0487} ±   0.0030 [1] &   0.0522 ±   0.0029 [3] \\
        shares & rnn &   0.3480 ±   0.0266  &   0.3435 ±   0.0142 [3] &   \textbf{0.2947} ±   0.0495 [1] \\
        shares &  cnn &   1.2505 ±   0.8296  &   0.3113 ±   0.0157 [3] &   \textbf{0.2153} ±   0.0175 [3] \\
          wine & rnn &   0.4097 ±   0.0203  &   0.3850 ±   0.0074 [3] &   \textbf{0.3432} ±   0.0043 [1] \\
          wine &  cnn &   1.3288 ±   0.5098  &   \textbf{0.3298} ±   0.0269 [2] &   0.4529 ±   0.0000 [2] \\
          
\bottomrule

\end{tabular}}
\label{table:denoise_vs_raw}
\caption{The table compares the best mean results obtained by applying denoising conditioned data augmentation with the mean performance of models trained using raw data. The number in brackets refers to the number of denoising steps applied.}
\end{table}

\subsection{Ablation study} \label{sec:ablation}

This section explores the effectiveness of denoising-based data augmentation, with and without meta-attribute conditioning, compared to data augmentation involving only noise introduction.

\subsubsection{Denoising process vs Noise augmentation}

\begin{figure}[t]
    \centering
    \includegraphics[width=1\textwidth]{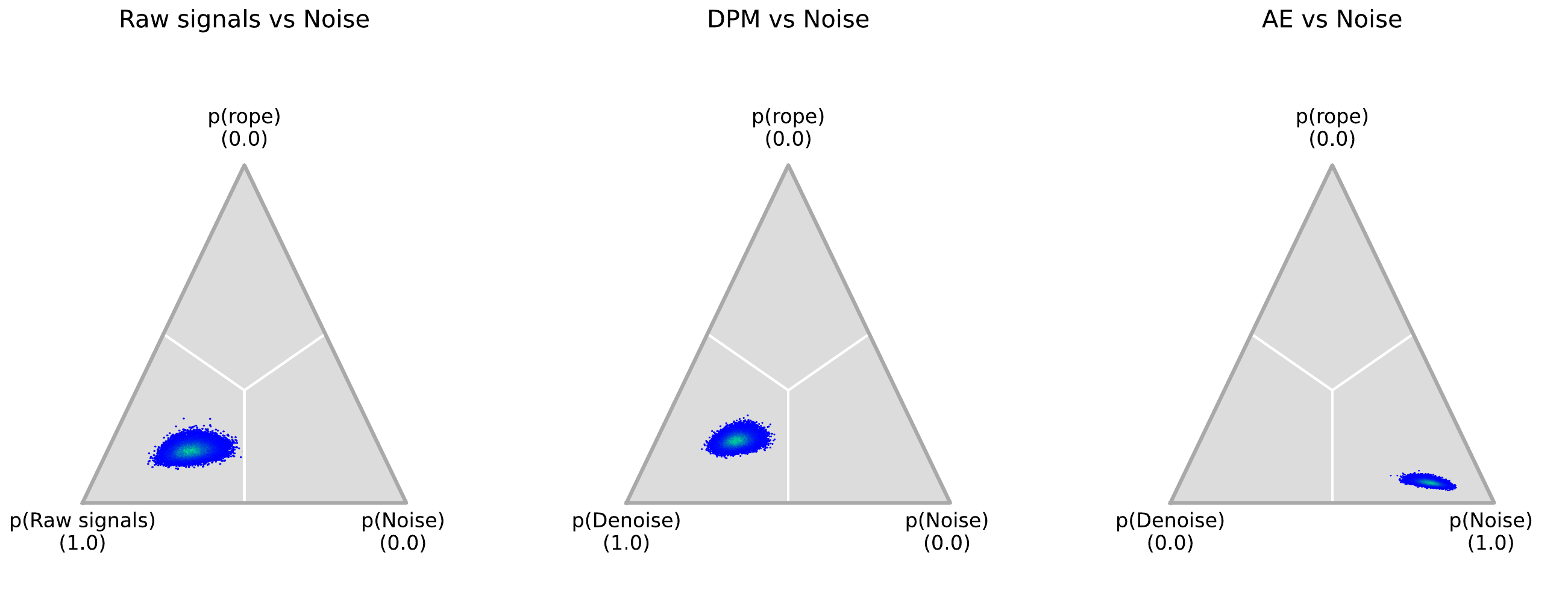}
    \caption{The chart on the left displays the comparative performance between conditioned noise-based data augmentation and the model trained solely with raw data. The center and right charts exhibit the performance obtained using denoising data augmentation based on DPM and autoencoders compared with that obtained using noise-based data augmentation.} 
    \label{fig:denoise_vs_noise}
\end{figure}

Figure \ref{fig:denoise_vs_noise} (left) displays the results of the Bayesian signed-rank test comparing conditioned noise-based data augmentation against models trained solely with raw data. As observed, noise data augmentation exhibits a low probability of generating a better generalized model when compared to models trained solely with raw data. Figure \ref{fig:denoise_vs_noise} (center and right) depicts the likelihood for the denoising-based data augmentation approaches (using DPM and autoencoders respectively) to perform better than the noise-based approach. Results distinctly demonstrates the superiority of DPM-based denoising models as data augmentation tool. %\red{Esta comparacion es con o sin conditioning}

%\red{Esta comparacion es con o sin conditioning}

\subsubsection{Meta-attribute conditioning}

\begin{figure}[t]
    \centering
    \includegraphics[width=1\textwidth]{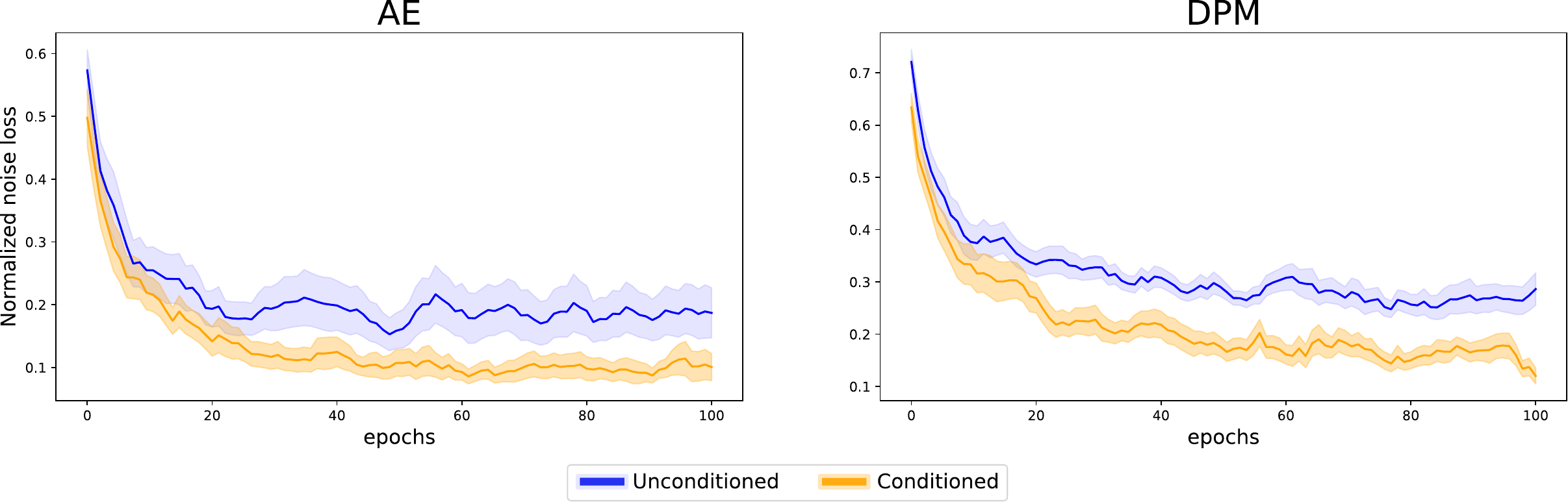}
    \caption{Normalized validation loss during the denoising model training process. The graph shows the confidence interval computed using the validation loss obtained for each dataset. Both graphs demonstrate that meta-attribute conditioning significantly contributes to the denoising learning process.}
    \label{fig:denoising_loss}
\end{figure}

Figure \ref{fig:denoising_loss} shows the training progress of the denoising models. To compute the confidence intervals and compare the contribution of the meta-attributes, the validation loss has been normalized per dataset. Results show that meta-attribute conditioning considerably improves the validation loss and reduces uncertainty.

\begin{figure}[t]
    \centering
    \includegraphics[width=0.8\textwidth]{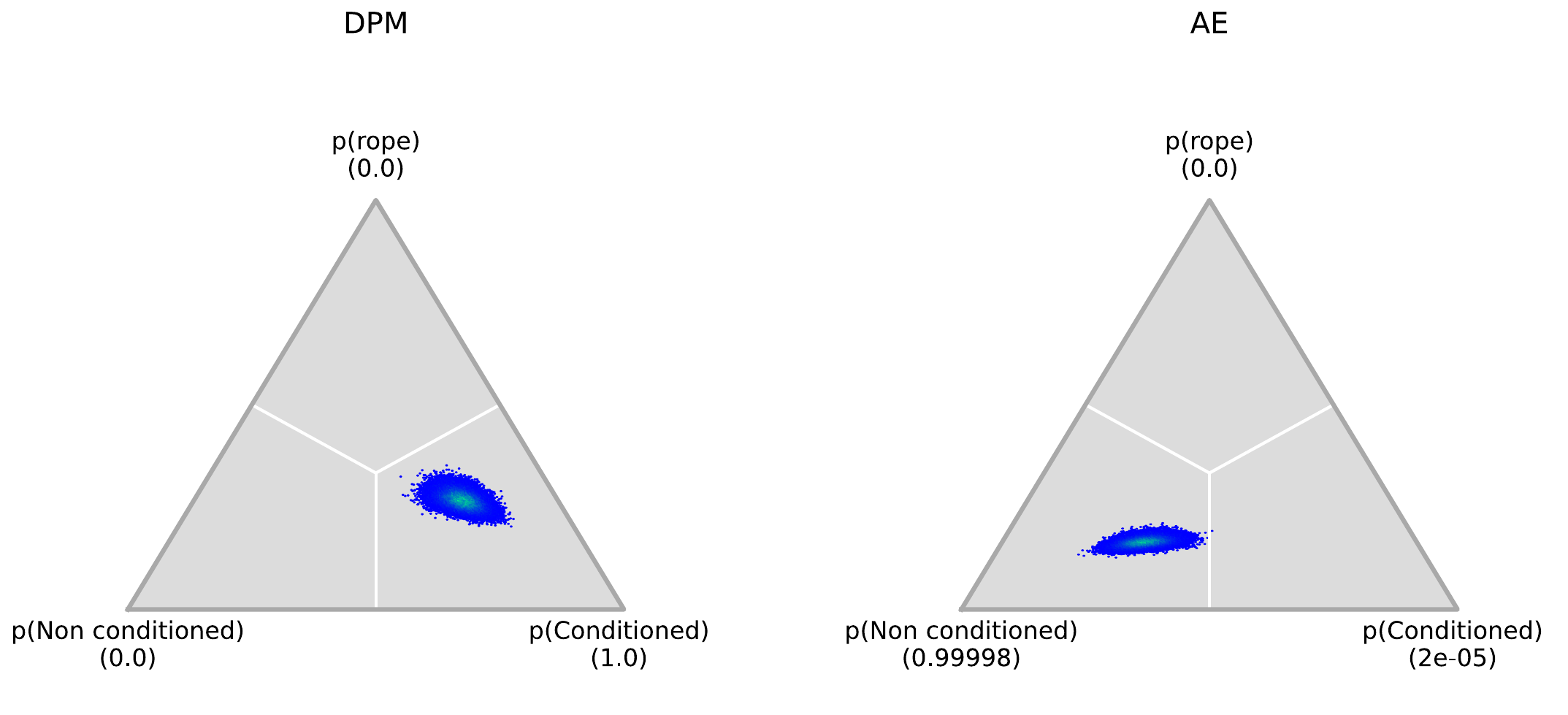}
    \caption{The chart depicts the influence of meta-attribute conditioning in the DPM and AE denoising data augmentation approaches. On the left, it is noticeable that conditioning substantialy impacts DPM approach. Conversely, the chart on the right demonstrates no impact of meta-attribute conditioning when AE are employed.}
    \label{fig:noncond_vs_cond}
\end{figure}

Figure \ref{fig:noncond_vs_cond} sheds light on the influence of meta-attribute conditioning when employing denoising-based data augmentation approaches. According to the results, it appears that the use of meta-attribute conditioning does not have a clear impact when combined with the AE models. However, when utilized alongside DPM models, it notably enhances the data augmentation process.

\begin{figure}[t]
    \centering
    \includegraphics[width=0.8\textwidth]{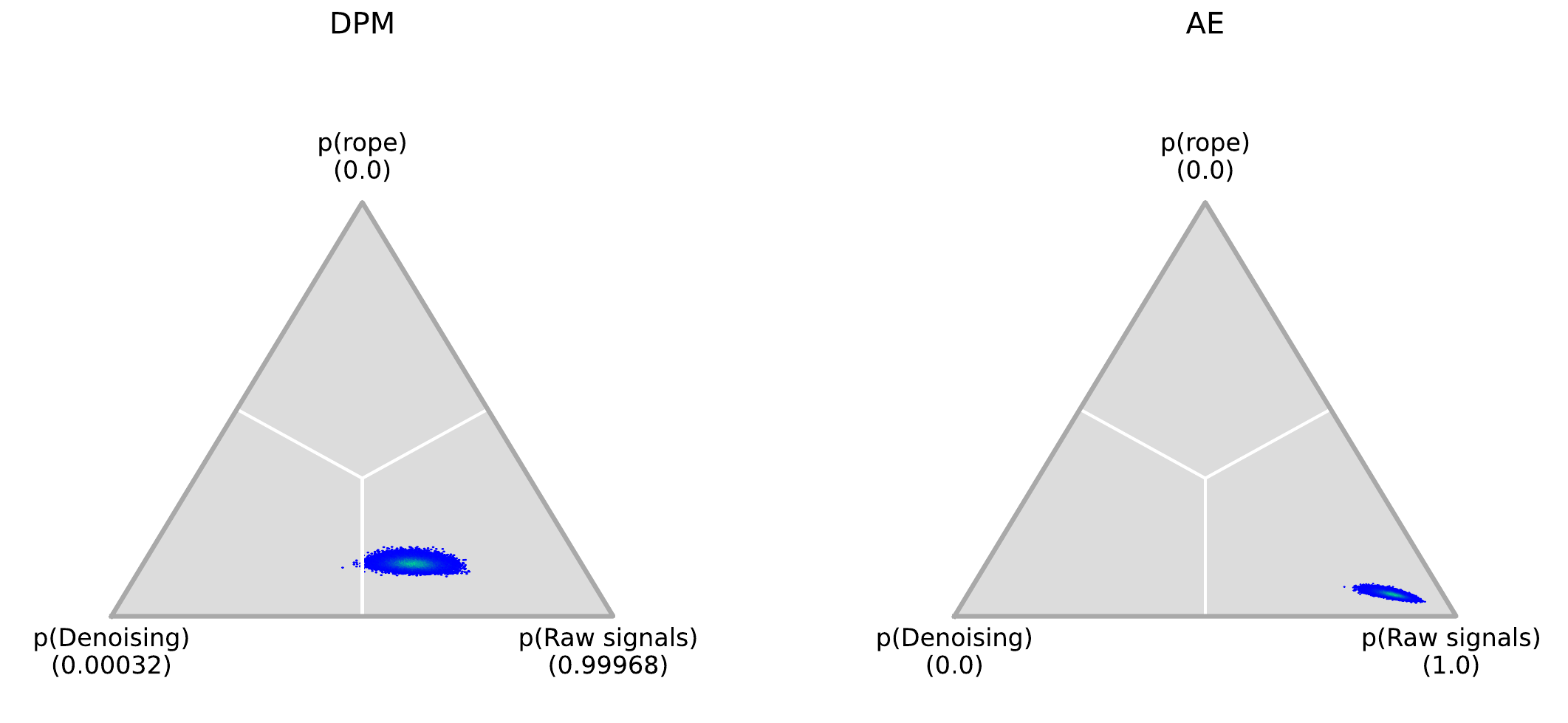}
    \caption{These charts depict the impact of unconditional denoising models compared to models trained solely with raw data. The comparison with Figure \ref{fig:denoise_vs_raw} highlights the influence of the conditioning, especially in DPM.}
    \label{fig:noncond_denoise_vs_raw}
\end{figure}

The lack of effectiveness in meta-attribute conditioning, when combined with the AE approach, might contribute to the superior performance of the DPM denoising approach. This hypothesis is supported when comparing non-conditioned approaches with raw data, as illustrated in Figure \ref{fig:noncond_denoise_vs_raw}. The removal of conditioning in the DPM drastically reduces the efficiency of the data augmentation process. However, the results of the AE remain very similar to those shown in Figure \ref{fig:denoise_vs_raw}.

\subsection{Number of denoising steps $T$ and noise rates $\beta_T$}

\begin{figure}[t]
    \centering
    \includegraphics[width=0.8\textwidth]{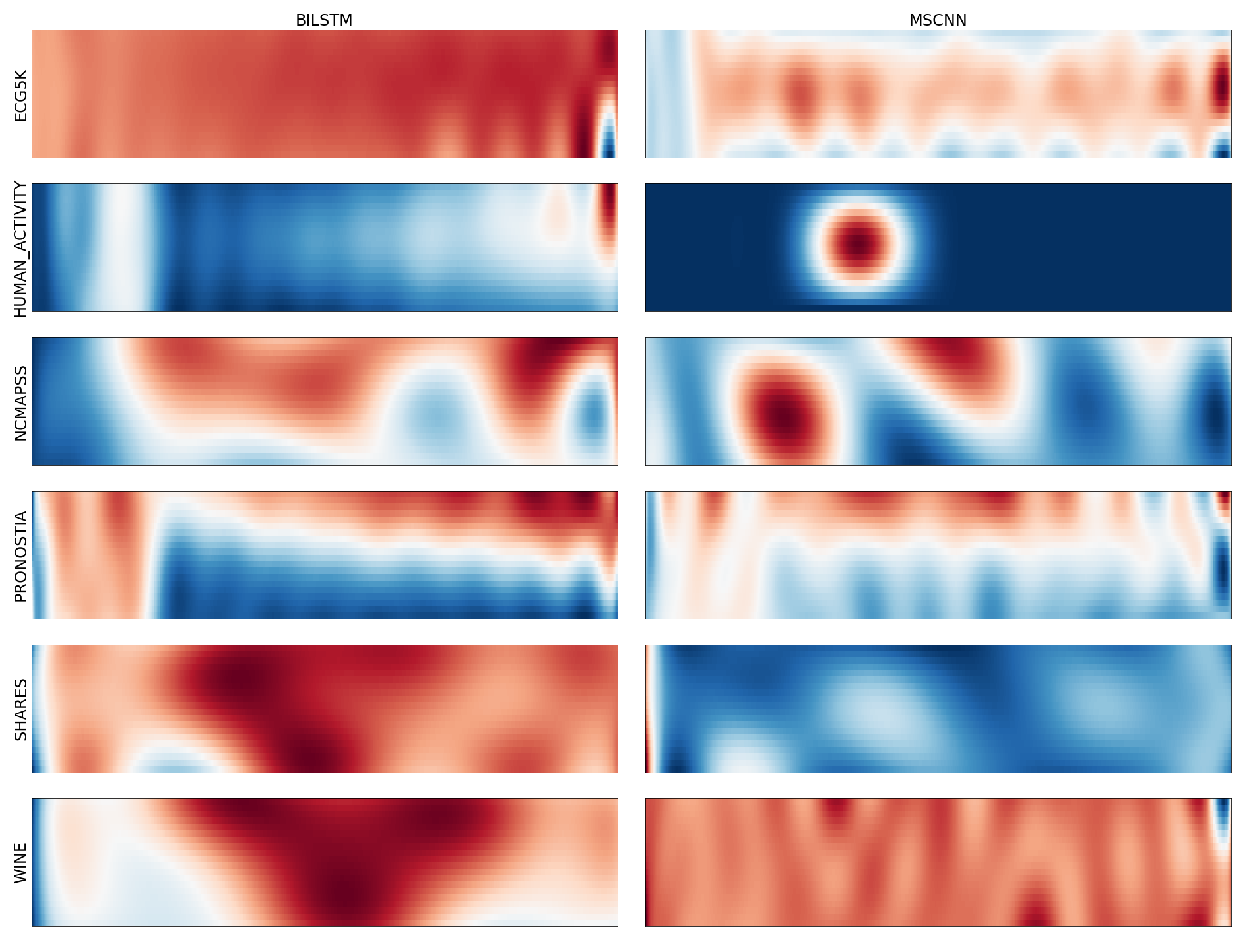}
    \caption{Comparison of the prediction performance using different denoising rates and denoising steps in denoising models. The y-axis of the charts corresponds to the number of denoising steps, and the x-axis corresponds to the noise rates added to the samples. Blue regions denote better performance than red regions.}
    \label{fig:steps_noise_rates}
\end{figure}

Figure \ref{fig:steps_noise_rates} depicts the search space of the hyperparameters $T$ (number of denoising steps) and $\beta_T$ (noise rate added to the original sample). The charts were created by fitting a Gaussian model with the results obtained during the experimentation, and creating an interpolation between the extreme values using the predictions of this Gaussian model. The analysis of this graph demonstrates that these hyperparameters have to be selected carefully, since the hyperparameter search space changes with each dataset and model architecture used. In general, the charts show that there are extensive areas where good configurations of these hyperparameters can be found.

\section{Conclusions and Future work}

This work has studied the application of denoising models for data augmentation in time series for classification and regression tasks. Additionally, a set of meta-attributes extracted from the raw data has been proposed to condition the network with the idea of enforcing the preservation of the sample category. Data expansion has been evaluated using single noise addition, denoising autoencoders (DAEs), and diffusion probabilistic models (DPMs) not previously studied in this area.

The study results have exhibited promising outcomes, demonstrating the potential of denoising models for effective data augmentation in diverse time series applications. Furthermore, the inclusion of meta-attributes has proven to be informative, enhancing the denoising process and providing a more effective data augmentation tool.

Beyond the promising results obtained, the proposed methodologies suffer some limitations. As a common issue previously demonstrated in autoencoders \cite{von2021self}, the introduced bias in the input will result in bias in the generated output. Thus, the denoising steps and noise rates need to be chosen carefully for the model to be applied successfully. Another important issue is the time requirements for denoising, which can make the training process slower. These issues will be the aim of a future work.

The proposed approaches have a excellent potential to be useful as data augmentation tools, but further work is required. The mentioned techniques introduce some hyperparameters (noise level and denoising steps) that either need optimization or exploration in some way. For instance, to avoid manual design, it might be interesting to learn how to apply an augmentation policy where different noises and denoising parameters are applied during the training process, similar to other works such as \cite{cubuk2019autoaugment} or \cite{cheung2021modals}. With the recent rise of large language models (LLM), it could be interesting to study whether this type of model, using the proper prompts, could help supporting the former task.

The meta-attributes have demonstrated to be sufficiently informative to enhance the denoising process. Additionally, they provide more descriptive information about the features of the sample. Therefore, a denoising model conditioned with the meta-attributes could be utilized to generate a neighborhood of a sample. This neighborhood could be employed in explainable artificial intelligence (XAI) methods such as LIME or SHAPLEY to explain the local prediction based on the meta-attributes variations. We believe that this idea could be useful in contexts like process optimization, predictive maintenance, or root cause analysis, where the application of XAI methods is more rare \cite{electronics12224572, solis2023soundness}.

In a similar fashion,  the process of generating new samples while preserving the statistical information supported by the meta-attributes could be employed in Differential Privacy (DP), as recent works explore in this direction \cite{chu2023differentially, he2023diff}.

\section*{Acknowledgments}

This work has been supported by Proyecto PID2019-109152GB-I00 financiado por MCIN/ AEI /10.13039/501100011033 (Agencia Estatal de Investigación), Spain and by the Ministry of Science and Education of Spain through the national program “Ayudas para contratos para la formación de investigadores en empresas (DIN2019-010887 / AEI / 10.13039/50110001103)”, of~State Programme of Science Research and Innovations 2017-2020.

\bibliographystyle{elsarticle-num} 
\bibliography{references}

\begin{thebibliography}{10}
\expandafter\ifx\csname url\endcsname\relax
  \def\url#1{\texttt{#1}}\fi
\expandafter\ifx\csname urlprefix\endcsname\relax\def\urlprefix{URL }\fi
\expandafter\ifx\csname href\endcsname\relax
  \def\href#1#2{#2} \def\path#1{#1}\fi

\bibitem{wen2020time}
Q.~Wen, L.~Sun, F.~Yang, X.~Song, J.~Gao, X.~Wang, H.~Xu, Time series data
  augmentation for deep learning: A survey, arXiv preprint arXiv:2002.12478
  (2020).

\bibitem{iglesias2022data}
G.~Iglesias, E.~Talavera, {\'A}.~Gonz{\'a}lez-Prieto, A.~Mozo,
  S.~G{\'o}mez-Canaval, Data augmentation techniques in time series domain: A
  survey and taxonomy, arXiv preprint arXiv:2206.13508 (2022).

\bibitem{liu2020efficient}
B.~Liu, Z.~Zhang, R.~Cui, Efficient time series augmentation methods, in: 2020
  13th international congress on image and signal processing, BioMedical
  engineering and informatics (CISP-BMEI), IEEE, 2020, pp. 1004--1009.

\bibitem{mi2022improving}
C.~Mi, L.~Xie, Y.~Zhang, Improving data augmentation for low resource
  speech-to-text translation with diverse paraphrasing, Neural Networks 148
  (2022) 194--205.

\bibitem{wang2018data}
F.~Wang, S.-h. Zhong, J.~Peng, J.~Jiang, Y.~Liu, Data augmentation for
  eeg-based emotion recognition with deep convolutional neural networks, in:
  MultiMedia Modeling: 24th International Conference, MMM 2018, Bangkok,
  Thailand, February 5-7, 2018, Proceedings, Part II 24, Springer, 2018, pp.
  82--93.

\bibitem{zur2009noise}
R.~M. Zur, Y.~Jiang, L.~L. Pesce, K.~Drukker, Noise injection for training
  artificial neural networks: A comparison with weight decay and early
  stopping, Medical physics 36~(10) (2009) 4810--4818.

\bibitem{moreno2018forward}
F.~J. Moreno-Barea, F.~Strazzera, J.~M. Jerez, D.~Urda, L.~Franco, Forward
  noise adjustment scheme for data augmentation, in: 2018 IEEE symposium series
  on computational intelligence (SSCI), IEEE, 2018, pp. 728--734.

\bibitem{xie2020unsupervised}
Q.~Xie, Z.~Dai, E.~Hovy, T.~Luong, Q.~Le, Unsupervised data augmentation for
  consistency training, Advances in neural information processing systems 33
  (2020) 6256--6268.

\bibitem{bishop1995training}
C.~M. Bishop, Training with noise is equivalent to tikhonov regularization,
  Neural computation 7~(1) (1995) 108--116.

\bibitem{goodfellow2014generative}
I.~Goodfellow, J.~Pouget-Abadie, M.~Mirza, B.~Xu, D.~Warde-Farley, S.~Ozair,
  A.~Courville, Y.~Bengio, Generative adversarial nets, Advances in neural
  information processing systems 27 (2014).

\bibitem{cabrera2019generative}
D.~Cabrera, F.~Sancho, J.~Long, R.-V. S{\'a}nchez, S.~Zhang, M.~Cerrada, C.~Li,
  Generative adversarial networks selection approach for extremely imbalanced
  fault diagnosis of reciprocating machinery, IEEE Access 7 (2019)
  70643--70653.

\bibitem{jenni2019stabilizing}
S.~Jenni, P.~Favaro, On stabilizing generative adversarial training with noise,
  in: Proceedings of the IEEE/CVF Conference on Computer Vision and Pattern
  Recognition, 2019, pp. 12145--12153.

\bibitem{roth2017stabilizing}
K.~Roth, A.~Lucchi, S.~Nowozin, T.~Hofmann, Stabilizing training of generative
  adversarial networks through regularization, Advances in neural information
  processing systems 30 (2017).

\bibitem{arjovsky2017towards}
M.~Arjovsky, L.~Bottou, Towards principled methods for training generative
  adversarial networks, arXiv preprint arXiv:1701.04862 (2017).

\bibitem{mogren2016c}
O.~Mogren, C-rnn-gan: Continuous recurrent neural networks with adversarial
  training, arXiv preprint arXiv:1611.09904 (2016).

\bibitem{esteban2017real}
C.~Esteban, S.~L. Hyland, G.~R{\"a}tsch, Real-valued (medical) time series
  generation with recurrent conditional gans, arXiv preprint arXiv:1706.02633
  (2017).

\bibitem{yoon2019time}
J.~Yoon, D.~Jarrett, M.~Van~der Schaar, Time-series generative adversarial
  networks, Advances in neural information processing systems 32 (2019).

\bibitem{lemley2017smart}
J.~Lemley, S.~Bazrafkan, P.~Corcoran, Smart augmentation learning an optimal
  data augmentation strategy, Ieee Access 5 (2017) 5858--5869.

\bibitem{rumelhart1985learning}
D.~E. Rumelhart, G.~E. Hinton, R.~J. Williams, et~al., Learning internal
  representations by error propagation (1985).

\bibitem{vincent2008extracting}
P.~Vincent, H.~Larochelle, Y.~Bengio, P.-A. Manzagol, Extracting and composing
  robust features with denoising autoencoders, in: Proceedings of the 25th
  international conference on Machine learning, 2008, pp. 1096--1103.

\bibitem{gondara2016medical}
L.~Gondara, Medical image denoising using convolutional denoising autoencoders,
  in: 2016 IEEE 16th international conference on data mining workshops (ICDMW),
  IEEE, 2016, pp. 241--246.

\bibitem{jiang2017wind}
G.~Jiang, P.~Xie, H.~He, J.~Yan, Wind turbine fault detection using a denoising
  autoencoder with temporal information, IEEE/Asme transactions on mechatronics
  23~(1) (2017) 89--100.

\bibitem{cabrera2017automatic}
D.~Cabrera, F.~Sancho, C.~Li, M.~Cerrada, R.-V. S{\'a}nchez, F.~Pacheco, J.~V.
  de~Oliveira, Automatic feature extraction of time-series applied to fault
  severity assessment of helical gearbox in stationary and non-stationary speed
  operation, Applied Soft Computing 58 (2017) 53--64.

\bibitem{rudolph2019structuring}
M.~Rudolph, B.~Wandt, B.~Rosenhahn, Structuring autoencoders, in: Proceedings
  of the IEEE/CVF International Conference on Computer Vision Workshops, 2019,
  pp. 0--0.

\bibitem{kingma2013auto}
D.~P. Kingma, M.~Welling, Auto-encoding variational bayes, arXiv preprint
  arXiv:1312.6114 (2013).

\bibitem{chadebec2021data}
C.~Chadebec, S.~Allassonni{\`e}re, Data augmentation with variational
  autoencoders and manifold sampling, in: Deep Generative Models, and Data
  Augmentation, Labelling, and Imperfections: First Workshop, DGM4MICCAI 2021,
  and First Workshop, DALI 2021, Held in Conjunction with MICCAI 2021,
  Strasbourg, France, October 1, 2021, Proceedings 1, Springer, 2021, pp.
  184--192.

\bibitem{islam2021crash}
Z.~Islam, M.~Abdel-Aty, Q.~Cai, J.~Yuan, Crash data augmentation using
  variational autoencoder, Accident Analysis \& Prevention 151 (2021) 105950.

\bibitem{nishizaki2017data}
H.~Nishizaki, Data augmentation and feature extraction using variational
  autoencoder for acoustic modeling, in: 2017 Asia-Pacific Signal and
  Information Processing Association Annual Summit and Conference (APSIPA ASC),
  IEEE, 2017, pp. 1222--1227.

\bibitem{momeny2021learning}
M.~Momeny, A.~A. Neshat, M.~A. Hussain, S.~Kia, M.~Marhamati, A.~Jahanbakhshi,
  G.~Hamarneh, Learning-to-augment strategy using noisy and denoised data:
  Improving generalizability of deep cnn for the detection of covid-19 in x-ray
  images, Computers in Biology and Medicine 136 (2021) 104704.

\bibitem{ho2020denoising}
J.~Ho, A.~Jain, P.~Abbeel, Denoising diffusion probabilistic models, Advances
  in neural information processing systems 33 (2020) 6840--6851.

\bibitem{rasul2021autoregressive}
K.~Rasul, C.~Seward, I.~Schuster, R.~Vollgraf, Autoregressive denoising
  diffusion models for multivariate probabilistic time series forecasting, in:
  International Conference on Machine Learning, PMLR, 2021, pp. 8857--8868.

\bibitem{li2022generative}
Y.~Li, X.~Lu, Y.~Wang, D.~Dou, Generative time series forecasting with
  diffusion, denoise, and disentanglement, Advances in Neural Information
  Processing Systems 35 (2022) 23009--23022.

\bibitem{goel2022s}
K.~Goel, A.~Gu, C.~Donahue, C.~R{\'e}, It’s raw! audio generation with
  state-space models, in: International Conference on Machine Learning, PMLR,
  2022, pp. 7616--7633.

\bibitem{chen2020wavegrad}
N.~Chen, Y.~Zhang, H.~Zen, R.~J. Weiss, M.~Norouzi, W.~Chan, Wavegrad:
  Estimating gradients for waveform generation, arXiv preprint arXiv:2009.00713
  (2020).

\bibitem{tashiro2021csdi}
Y.~Tashiro, J.~Song, Y.~Song, S.~Ermon, Csdi: Conditional score-based diffusion
  models for probabilistic time series imputation, Advances in Neural
  Information Processing Systems 34 (2021) 24804--24816.

\bibitem{christ2018time}
M.~Christ, N.~Braun, J.~Neuffer, A.~W. Kempa-Liehr, Time series feature
  extraction on basis of scalable hypothesis tests (tsfresh--a python package),
  Neurocomputing 307 (2018) 72--77.

\bibitem{ref-arias}
M.~Arias~Chao, C.~Kulkarni, K.~Goebel, O.~Fink, Aircraft engine run-to-failure
  dataset under real flight conditions for prognostics and diagnostics, Data
  6~(1) (2021) 5.

\bibitem{solis2021stacked}
D.~Solis-Martin, J.~Gal{\'a}n-P{\'a}ez, J.~Borrego-Diaz, A stacked deep
  convolutional neural network to predict the remaining useful life of a
  turbofan engine, arXiv preprint arXiv:2111.12689 (2021).

\bibitem{solis4596645deep}
D.~Sol{\'\i}s-Mart{\'\i}n, J.~Galan-Paez, J.~D{\'\i}az, A deep learning
  framework to predict the remaining useful life in predictive maintenance,
  Available at SSRN 4596645.

\bibitem{cohen2023shapley}
J.~Cohen, X.~Huan, J.~Ni, Shapley-based explainable ai for clustering
  applications in fault diagnosis and prognosis, arXiv preprint
  arXiv:2303.14581 (2023).

\bibitem{ref-pronostia}
P.~Nectoux, R.~Gouriveau, K.~Medjaher, E.~Ramasso, B.~Chebel-Morello,
  N.~Zerhouni, C.~Varnier, Pronostia: An experimental platform for bearings
  accelerated degradation tests., in: IEEE International Conference on
  Prognostics and Health Management, PHM'12., IEEE Catalog Number:
  CPF12PHM-CDR, 2012, pp. 1--8.

\bibitem{goldberger2000physiobank}
A.~L. Goldberger, L.~A. Amaral, L.~Glass, J.~M. Hausdorff, P.~C. Ivanov, R.~G.
  Mark, J.~E. Mietus, G.~B. Moody, C.-K. Peng, H.~E. Stanley, Physiobank,
  physiotoolkit, and physionet: components of a new research resource for
  complex physiologic signals, circulation 101~(23) (2000) e215--e220.

\bibitem{anguita2013public}
D.~Anguita, A.~Ghio, L.~Oneto, X.~Parra, J.~L. Reyes-Ortiz, et~al., A public
  domain dataset for human activity recognition using smartphones., in: Esann,
  Vol.~3, 2013, p.~3.

\bibitem{timeseriesclassification}
A.~Bagnall, J.~Lines, W.~Vickers, E.~Keogh,
  \href{www.timeseriesclassification.com}{Time series classification
  repository} (2021).
\newline\urlprefix\url{www.timeseriesclassification.com}

\bibitem{cui2016multi}
Z.~Cui, W.~Chen, Y.~Chen, Multi-scale convolutional neural networks for time
  series classification, arXiv preprint arXiv:1603.06995 (2016).

\bibitem{benavoli2017time}
A.~Benavoli, G.~Corani, J.~Dem{\v{s}}ar, M.~Zaffalon, Time for a change: a
  tutorial for comparing multiple classifiers through bayesian analysis, The
  Journal of Machine Learning Research 18~(1) (2017) 2653--2688.

\bibitem{benavoli2014bayesian}
A.~Benavoli, G.~Corani, F.~Mangili, M.~Zaffalon, F.~Ruggeri, A bayesian
  wilcoxon signed-rank test based on the dirichlet process, in: International
  conference on machine learning, PMLR, 2014, pp. 1026--1034.

\bibitem{von2021self}
J.~Von~K{\"u}gelgen, Y.~Sharma, L.~Gresele, W.~Brendel, B.~Sch{\"o}lkopf,
  M.~Besserve, F.~Locatello, Self-supervised learning with data augmentations
  provably isolates content from style, Advances in neural information
  processing systems 34 (2021) 16451--16467.

\bibitem{cubuk2019autoaugment}
E.~D. Cubuk, B.~Zoph, D.~Mane, V.~Vasudevan, Q.~V. Le, Autoaugment: Learning
  augmentation strategies from data, in: Proceedings of the IEEE/CVF conference
  on computer vision and pattern recognition, 2019, pp. 113--123.

\bibitem{cheung2021modals}
T.-H. Cheung, D.-Y. Yeung, Modals: Modality-agnostic automated data
  augmentation in the latent space, in: International Conference on Learning
  Representations, 2021.

\bibitem{electronics12224572}
R.~Hoffmann, C.~Reich, \href{https://www.mdpi.com/2079-9292/12/22/4572}{A
  systematic literature review on artificial intelligence and explainable
  artificial intelligence for visual quality assurance in manufacturing},
  Electronics 12~(22) (2023).
\newblock \href {https://doi.org/10.3390/electronics12224572}
  {\path{doi:10.3390/electronics12224572}}.
\newline\urlprefix\url{https://www.mdpi.com/2079-9292/12/22/4572}

\bibitem{solis2023soundness}
D.~Sol{\'\i}s-Mart{\'\i}n, J.~Gal{\'a}n-P{\'a}ez, J.~Borrego-D{\'\i}az, On the
  soundness of xai in prognostics and health management (phm), Information
  14~(5) (2023) 256.

\bibitem{chu2023differentially}
Z.~Chu, J.~He, D.~Peng, X.~Zhang, N.~Zhu, Differentially private denoise
  diffusion probability models, IEEE Access (2023).

\bibitem{he2023diff}
X.~He, M.~Zhu, D.~Chen, N.~Wang, X.~Gao, Diff-privacy: Diffusion-based face
  privacy protection, arXiv preprint arXiv:2309.05330 (2023).

\end{thebibliography}

\end{document}